\documentclass[10pt, a4paper]{article}
\usepackage{lrec}
\usepackage[utf8]{inputenc}
\usepackage{graphicx}
\usepackage{tabularx}
\usepackage{soul}
\usepackage{enumitem}
\usepackage{epstopdf}
\usepackage{xcolor}
\usepackage{hyperref}
\usepackage{xstring}
\usepackage{listings}
\usepackage[toc,page]{appendix}

\title{Introducing RONEC - the Romanian Named Entity Corpus}

\name{Stefan Daniel Dumitrescu, Andrei-Marius Avram\textsuperscript{1}}

\address{University Politehnica of Bucharest\textsuperscript{1} \\
         Bucharest, Romania \\
        dumitrescu.stefan@gmail.com, avram.andreimarius@gmail.com\\}

\abstract{
We present RONEC - the Named Entity Corpus for the Romanian language. The corpus contains over 26000 entities in ~5000 annotated sentences, belonging to 16 distinct classes. The sentences have been extracted from a copy-right free newspaper, covering several styles. This corpus represents the first initiative in the Romanian language space specifically targeted for named entity recognition. It is available as BRAT and CoNLL-U Plus (in Multi-Word Expression and IOB formats) text downloads, and it is free to use and extend at github.com/dumitrescustefan/ronec . \\
\newline  }

\begin{document}

\maketitleabstract

\section{Introduction}

Language resources are an essential component in entire R\&D domains. From the humble but vast repositories of monolingual texts that are used by the newest language modeling approaches like BERT\footnote{BERT, released in 2018, is the baseline for today's much more advanced systems.} and GPT\footnote{OpenAI's GPT-2 \cite{radford2019language} is a very strong text generation model.}, to parallel corpora that allows our machine translation systems to inch closer to human performance, to the more specialized resources like WordNets\footnote{RoWordNet \cite{dumitrescu2018rowordnet} is a relatively recent resource in the Romanian language space.} that encode semantic relations between nodes, these resources are necessary for the general advancement of Natural Language Processing, which eventually evolves into real apps and services we are (already) taking for granted.

We introduce \textbf{RONEC} - the \textbf{RO}manian \textbf{N}amed \textbf{E}ntity \textbf{C}orpus\footnote{RONEC ISLRN: 723-333-596-623-8, available at \url{https://github.com/dumitrescustefan/ronec} }, a free, open-source resource that contains annotated named entities in copy-right free text.

A named entity corpus is generally used for Named Entity Recognition (NER): the identification of entities in text such as names of persons, locations, companies, dates, quantities, monetary values, etc. This information would be very useful for any number of applications: from a general information extraction system down to task-specific apps such as identifying monetary values in invoices or product and company references in customer reviews.

We motivate \textit{the need} for this corpus primarily because, for Romanian, there is \textit{no other such corpus.} This basic necessity has sharply arisen as we, while working on a different project, have found out there are no usable resources to help us in an Information Extraction task: we were unable to extract people, locations or dates/values. This constituted a major road-block, with the only solution being to create such a corpus ourselves. As the corpus was out-of-scope for this project, the work was done privately, outside the umbrella of any authors' affiliations - this is why we are able to distribute this corpus completely free\footnote{Unfortunately, many Romanian language resources have been developed in different funded projects and carry stronger copyright licenses, including requiring potential users to print/ sign/scan/send copyright forms, a step that discourages the vast majority of people.}.

The current landscape in Romania regarding language resources is relatively unchanged from the outline given by the META-NET\footnote{META-NET  website: \url{http://www.meta-net.eu/}} project over six years ago. The in-depth analysis performed in this European-wide Horizon2020-funded project revealed that the Romanian language falls in the "fragmentary support" category, just above the last, "weak/none" category (see the language/support matrix in \cite{rehm2013meta}). This is why, in 2019/2020, we are able to present the first Romanian NER resource.

\section{Related Work}

We note that, while fragmentary, there are a few related language resources available, but \textit{none} that \textit{specifically} target named entities:

\subsection{ROCO corpus}
ROCO\footnote{ROCO ISLRN: 312-617-089-348-7, ELRA-W0085} is a Romanian journalistic corpus that contains approx. 7.1M tokens. It is rich in proper names, numerals and named entities. The corpus has been automatically annotated at word-level with morphosyntactic information (MSD annotations).

\subsection{ROMBAC corpus}
Released in 2016, ROMBAC\footnote{ROMBAC ISLRN: 162-192-982-061-0, ELRA-W0088} is a Romanian text corpus containing ~41M words divided in relatively equal domains like journalism, legalese, fiction, medicine, etc. Similarly to ROCO, it is automatically annotated at word level with MSD descriptors.

\subsection{CoRoLa corpus}
The much larger and recently released CoRoLa corpus\footnote{CoRoLa available at: \url{http://corola.racai.ro/}} contains over 1B words, similarly automatically annotated. 
\medskip

In all these corpora the named entities are not a separate category - the texts are morphologically and syntactically annotated and all proper nouns are marked as such - NP - without any other annotation or assigned category. Thus, these corpora cannot be used in a true NER sense. Furthermore, annotations were done automatically with a tokenizer/tagger/parser, and thus are of slightly lower quality than one would expect of a gold-standard corpus.

\section{Corpus Description}

The corpus, at its current version 1.0 is composed of \textbf{5127 sentences}, annotated with \textbf{16 classes}, for a total of \textbf{26377 annotated entities}. The 16 classes are: PERSON, NAT\_REL\_POL, ORG, GPE, LOC, FACILITY, PRODUCT, EVENT, LANGUAGE, WORK\_OF\_ART, DATETIME, PERIOD, MONEY, QUANTITY, NUMERIC\_VALUE and ORDINAL.

It is based on copyright-free text extracted from Southeast European Times (SETimes) \cite{tyers2010south}. The news portal has published\footnote{setimes.com has ended publication in March 2015} “news and views from Southeast Europe” in ten languages, including Romanian. SETimes has been used in the past for several annotated corpora, including parallel corpora for machine translation. For RONEC we have used a hand-picked\footnote{We tried to select sentences so as to both maximize the amount of named entities while also keep a balanced domain coverage.} selection of sentences belonging to several categories (see table \ref{tab:styleexamples} for stylistic examples). 

\begin{table*}[ht]
\begin{center}
\begingroup
\renewcommand{\arraystretch}{1.3}
\begin{tabular}{|m{3.5cm}|m{12.8cm}|}

   \hline
   Style & Example sentence\\
   \hline\hline
   Current news & În \textbf{două zile}, luptele de la \textbf{Fallujah} din \textbf{Irak} au provocat moartea a \textbf{105 persoane} și rănirea a peste alte \textbf{200}.\\
   Historical news & \textbf{Jean-Claude Juncker}, premierul \textbf{luxemburghez} s-a născut în \textbf{9 decembrie 1954}.\\
   Free time & \textbf{Turiștii} care doresc să-și petreacă vacanța într-un loc liniștit, frumos și cu minim de cheltuieli pot opta pentru spațiile special amenajate pentru \textbf{corturi} atât la munte, cât și la mare sau în \textbf{Delta Dunării}.\\
   Sports & Tot în cadrul etapei \textbf{a 2-a}, a avut loc întâlnirea \textbf{Vardar Skopje} - \textbf{S.C. Pick Szeged}, care s-a încheiat la egalitate, \textbf{24} - \textbf{24}.\\
   Juridical news pieces& Ordonanța Guvernului nr. \textbf{83} / \textbf{2004} pentru modificarea și completarea Legii nr. \textbf{57} / \textbf{2003} privind Codul fiscal prevede, la art. \textbf{253}, alineatul (\textbf{6})... \\
   Personal adverts (e.g. buying-selling) & \textbf{S.C. "Innuendo” S.R.L.} vinde en gros, prin intermediul depozitului propriu situat în incinta \textbf{Centrului Comercial "Euro 1”}...\\
   Editorials (written sometimes in first person) & Pe \textbf{Valea Cernei} am ajuns, de această dată, pe drumul (\textbf{DN67D}) dinspre \textbf{Baia de Aramă}.\\
   \hline

\end{tabular}
\endgroup
\caption{Stylistic domains and examples (bold marks annotated entities). Translations are depicted in Appendix A.}
\label{tab:styleexamples}
\end{center}
\end{table*}

The corpus contains the standard diacritics in Romanian: letters \textbf{ș} and \textbf{ț} are written with a comma, not with a cedilla (like \textbf{ş} and \textbf{ţ}). In Romanian many older texts are written with cedillas instead of commas because full Unicode support in Windows came much later than the classic extended ASCII which only contained the cedilla letters.

The 16 classes are inspired by the OntoNotes5 corpus \cite{weischedel2013ontonotes} as well as the ACE (Automatic Content Extraction) English Annotation Guidelines for Entities Version 6.6 2008.06.13 \cite{linguistic2005ace}. We dropped 2 classes from OntoNote's 18 classes\footnote{Compared to OntoNotes we dropped its LAW class as it had almost no entity in our corpus, and compressed DATE and TIME into DATETIME, as surprisingly we found many cases where the distinction between DATE and TIME would be confusing for annotators. Furthermore, DATETIME entities will usually require further sub-processing to extract exact values, something which is out of scope for this corpus.}. Each one will be presented in detail, in section \ref{methodology} A summary of available classes with word counts for each is available in table \ref{tab:stats}.

\begin{table}[hbt!]
\centering
\begin{tabular}{|lccc|}
\hline
Class   & Total  & \textbf{Total}  & Words \\
        & words & \textbf{entities} & per entity \\
\hline\hline
PERSON & 10251 & 5363 & 1.911 \\
NAT\_REL\_POL & 1353 & 1324 & 1.022 \\
ORGANIZATION & 9794 & 3410 & 2.872 \\
GPE & 4751 & 4180 & 1.137 \\
LOC & 2364 & 920 & 2.57 \\
FACILITY & 2510 & 1187 & 2.115 \\
PRODUCT & 2042 & 1331 & 1.534 \\
EVENT & 1341 & 425 & 3.155 \\
LANGUAGE & 98 & 97 & 1.01 \\
WORK\_OF\_ART & 863 & 248 & 3.48 \\
DATETIME & 7072 & 3003 & 2.355 \\
PERIOD & 1295 & 385 & 3.364 \\
MONEY & 2591 & 898 & 2.885 \\
QUANTITY & 769 & 360 & 2.136 \\
NUMERIC\_VALUE & 2807 & 2714 & 1.034 \\
ORDINAL & 859 & 532 & 1.615 \\
\hline\hline
Total & 50760 & \textbf{26377} & 2.137 \\
\hline

\end{tabular}
\caption{Corpus statistics: Each entity is marked with a class and can span one or more words}
\label{tab:stats}
\end{table}

The corpus is available in two formats: \textbf{BRAT} and \textbf{CoNLL-U Plus} (MWE and IOB styles). 

\subsection{BRAT format}

As the corpus was developed in the BRAT \footnote{BRAT Rapid Annotation Tool:  \url{http://brat.nlplab.org/} } environment, it was natural to keep this format as-is. BRAT is an online environment for collaborative text annotation - a web-based tool where several people can mark words, sub-word pieces, multiple word expressions, can link them together by relations, etc. The back-end format is very simple: given a text file that contains raw sentences, in another text file every annotated entity is specified by the start/end character offset as well as the entity type, one per line. 
RONEC is exported in the BRAT format as ready-to-use in the BRAT annotator itself. The corpus is pre-split into sub-folders, and contains all the extra files such as the entity list, etc, needed to directly start an eventual edit/extension of the corpus. 

Example (raw/untokenized) sentences:
{\footnotesize \fontfamily{qcr}\selectfont

Tot în cadrul etapei \textbf{a 2-a}, a avut loc întâlnirea \textbf{Vardar Skopje} - \textbf{S.C. Pick Szeged}, care s-a încheiat la egalitate, \textbf{24} - \textbf{24}.

I s-a decernat Premiul Nobel pentru literatură pe \textbf{anul 1959}.
}

Example annotation format:

{\footnotesize \fontfamily{qcr}\selectfont
T1	ORDINAL 21 26	\textbf{a 2-a}

T2	ORGANIZATION 50 63	\textbf{Vardar Skopje}

T3	ORGANIZATION 66 82	\textbf{S.C. Pick Szeged}

T4	NUMERIC\_VALUE 116 118 \textbf{24}

T5	NUMERIC\_VALUE 121 123 \textbf{24}

T6	DATETIME 175 184	\textbf{anul 1959}
}

\subsection{CoNLL-U Plus format}

The CoNLL-U Plus\footnote{CoNLL-U Plus format description available at: \url{http://universaldependencies.org/ext-format.html}} format extends the standard CoNLL-U which is used to annotate sentences, and in which many corpora are found today. The CoNLL-U format annotates one word per line with 10 distinct "columns" (tab separated): 
\smallskip
\setlist{nolistsep}
\begin{enumerate}[noitemsep]
    \item ID: word index; 
    \item FORM: unmodified word from the sentence;
    \item LEMMA: the word's lemma;
    \item UPOS: Universal part-of-speech tag;
    \item XPOS: Language-specific part-of-speech tag;
    \item FEATS: List of morphological features from the universal feature inventory or from a defined language-specific extension;
    \item HEAD: Head of the current word, which is either a value of ID or zero;
    \item DEPREL: Universal dependency relation to the HEAD or a defined language-specific subtype of one;
    \item DEPS: Enhanced dependency graph in the form of a list of head-deprel pairs;
    \item MISC: Miscellaneous annotations such as space after word.
\end{enumerate}
\medskip

The CoNLL-U Plus extends this format by allowing a variable number of columns, with the restriction that the columns are to be defined in the header. For RONEC, we define our CoNLL-U Plus format as the standard 10 columns \textbf{plus another extra column named RONEC:CLASS}. This column has the following format\footnote{based on the PARSEME:MWE multi-word expressions, see  \href{http://multiword.sourceforge.net/PHITE.php?sitesig=CONF&page=CONF_04_LAW-MWE-CxG_2018___lb__COLING__rb__&subpage=CONF_45_Format_specification}{CUPT format here}. }: 
\setlist{nolistsep}
\begin{itemize}[noitemsep]
    \item each named entity has a distinct id in the sentence, starting from 1; as an entity can span several words, all words that belong to it have the same id (no relation to word indexes)
    \item the first word belonging to an entity also contains its class (e.g. word "John" in entity "John Smith" will be marked as "1:PERSON")
    \item a non-entity word is marked with an asterisk *
\end{itemize}
\medskip

Table \ref{tab:conllupformat} shows the CoNLL-U Plus format where for example "a 2-a" is an ORDINAL entity spanning 3 words. The first word "a" is marked in this last column as "1:ORDINAL" while the following words just with the id "1".

We also release the same CoNLL-U Plus format but the last column is encoded in the classic IOB format. 

\begin{table*}[ht]
\begin{center}
\begingroup
\renewcommand{\arraystretch}{1.0}
\begin{tabular}{|llllllll|}
   \hline
   ID & FORM & LEMMA & UPOS & XPOS & HEAD & DEPREL & \textbf{RONEC:CLASS} \\
   (\#1) &  (\#2) &  (\#3) &  (\#4) &  (\#5) &  (\#7) &  (\#8) &  \textbf{(\#11)} \\
   \hline\hline
1	&	Tot	&	tot	&	ADV	&	Rp	                &	3	&	advmod	&	*\\
2	&	în	&	în	&	ADP	&	Spsa	            &	3	&	case	&	*\\
3	&	cadrul	&	cadru	&	NOUN	  &	Ncmsry	&	10	&	obl	    &	*\\
4	&	etapei	&	etapă	&	NOUN	  &	Ncfsoy	&	3	&	nmod	&	*\\
5	&	\textbf{a}	&	al	&	DET	&	Tsfs	            &	6	&	det	    &	\textbf{1:ORDINAL}\\
6	&	\textbf{2}	&	2	&	NUM	&	Mc-p-d	            &	4	&	nummod	&	\textbf{1}\\
7	&	\textbf{-a}	&	-a	&	DET	&	Tffs-y	            &	6	&	det	    &	\textbf{1}\\
8	&	,	&	,	&	PUNCT	&	COMMA	        &	3	&	punct	&	*\\
9	&	a	&	avea	&	AUX	&	Va--3s	        &	10	&	aux	    &	*\\
10	&	avut	&	avea	&	VERB	&	Vmp--sm	&	0	&	root	&	*\\
11	&	loc	&	loc	&	NOUN	&	Ncms-n	        &	10	&	fixed	&	*\\
12	&	întâlnirea&întâlnire&NOUN&Ncfsry            &   10  &   nsubj	&	*\\
13	&	\textbf{Vardar}	&	Vardar	&	PROPN	&	Np	    &	12	&	nmod	&	\textbf{2:ORGANIZATION}\\
14	&	\textbf{Skopje}	&	Skopje	&	PROPN	&	Np	    &	13	&	flat	&	\textbf{2}\\
15	&	-	&	-	&	PUNCT	&	DASH	        &	13	&	punct	&	*\\
16	&	\textbf{S.C.}	&	s.c.	&	NOUN	&	Yn	    &	13	&	conj	&	\textbf{3:ORGANIZATION}\\
17	&	\textbf{Pick}	&	Pick	&	PROPN	&	Np	    &	13	&	flat	&	\textbf{3}\\
18	&	\textbf{Szeged}	&	Szeged	&	PROPN	&	Np	    &	17	&	flat	&	\textbf{3}\\
19	&	,	&	,	&	PUNCT	&	COMMA	        &	23	&	punct	&	*\\
20	&	care	&	care	&	PRON	&	Pw3--r	&	23	&	nsubj	&	*\\

   \hline

\end{tabular}
\endgroup
\caption{CoNLL-U Plus format for the first 20 tokens of sentence "\textit{Tot în cadrul etapei a 2-a, a avut loc întâlnirea Vardar Skopje - S.C. Pick Szeged, care s-a încheiat la egalitate, 24 - 24.}" (bold marks entities). The format is a text file containing a token per line annotated with 11 tab-separated columns, with an empty line marking the start of a new sentence. Please note that only column \#11 is human annotated (and the target of this work), the rest of the morpho-syntactic annotations have been automatically generated with NLP-Cube.}
\label{tab:conllupformat}
\end{center}
\end{table*}

The CoNLL-U Plus format we provide was created as follows: (1) annotate the raw sentences using the  NLP-Cube\footnote{NLP-Cube is a multilingual text preprocessing tool with SOTA-level accuracy, that exports directly in CoNLL format and is available at \url{https://github.com/adobe/NLP-Cube} } \cite{boros2018nlp} tool for Romanian (it provides everything from tokenization to parsing, filling in all attributes in columns \#1-\#10); (2) align each token with the human-made entity annotations from the BRAT environment (the alignment is done automatically and is error-free) and fill in column \#11.

\section{RONEC Classes}
\label{methodology}

For the English language, we found two "categories" of NER annotations to be more prominent: CoNLL- and ACE-style. Because CoNLL only annotates a few classes (depending on the corpora, starting from the basic three: PERSON, ORGANIZATION and LOCATION, up to seven), we chose to follow the ACE-style with 18 different classes. After analyzing the ACE guide we have settled on 16 final classes that seemed more appropriate for Romanian, seen in table \ref{tab:stats}.

In the following sub-sections we will describe each class in turn, with a few examples. Some examples have been left in Romanian while some have been translated in English for the reader's convenience. In the examples at the end of each class' description, translations in English are colored for easier reading.

\subsection{PERSON}

Persons, including fictive characters. We also mark common nouns that refer to a person (or several), including pronouns (us, them, they), but not articles (e.g. in "an individual" we don't mark "an"). Positions are not marked unless they directly refer to the person: "The presidential counselor has advised ... that a new counselor position is open.", here we mark "presidential counselor" because it refers to a person and not the "counselor" at the end of the sentence as it refers only to a position.

\medskip \hrule \medskip 

{\footnotesize \fontfamily{qcr}\selectfont
Locul doi i-a revenit româncei \textbf{Otilia Aionesei}, o \textbf{elevă} de 17 ani.

\textcolor{green!55!blue}{The second place was won by \textbf{Otilia Aionesei}, a 17 year old \textbf{student}.}
\medskip \hrule \medskip 

\textbf{Ministrul} bulgar \textbf{pentru afaceri europene, Meglena Kuneva} ...

\textcolor{green!55!blue}{The Bulgarian \textbf{Minister for European Affairs}, \textbf{Meglena Kuneva} ...\footnote{Note: in Romanian word ordering makes for two entities while in English it looks like just one.}}

}

\medskip \hrule \medskip 

\subsection{NAT\_REL\_POL}

These are nationalities or religious or political groups. We include words that indicate the nationality of a person, group or product/object. Generally words marked as NAT\_REl\_POL are adjectives.

\medskip \hrule \medskip 

{\footnotesize \fontfamily{qcr}\selectfont
avionul \textbf{american}

\textcolor{green!55!blue}{the \textbf{American} airplane}
\medskip \hrule \medskip 
Grupul \textbf{olandez}

\textcolor{green!55!blue}{the \textbf{Dutch} group}
\medskip \hrule \medskip 

\textbf{Grecii} iși vor alege președintele.

\textcolor{green!55!blue}{The \textbf{Greeks} will elect their president.}

}
\medskip \hrule \medskip 

\subsection{ORGANIZATION}

Companies, agencies, institutions, sports teams, groups of people. These entities must have an organizational structure. We only mark full organizational entities, not fragments, divisions or sub-structures.

\medskip \hrule \medskip 

{\footnotesize \fontfamily{qcr}\selectfont
\textbf{Universitatea Politehnica București} a decis ...

\textcolor{green!55!blue}{The \textbf{Politehnic University of Bucharest} has decided ...}

\medskip \hrule \medskip 

\textbf{Adobe Inc.} a lansat un nou produs.

\textcolor{green!55!blue}{\textbf{Adobe Inc.} has launched a new product.}
}
\medskip \hrule \medskip 

\subsection{GPE}

Geo-political entities: countries, counties, cities, villages. GPE entities have \textit{all} of the following components: (1) a population, (2) a well-defined governing/organizing structure and (3) a physical location. GPE entities are not sub-entities (like a neighbourhood from a city).
\medskip \hrule \medskip 

{\footnotesize \fontfamily{qcr}\selectfont
Armin van Buuren s-a născut în \textbf{Leiden}.

\textcolor{green!55!blue}{Armin van Buuren was born in \textbf{Leiden}.}

\medskip \hrule \medskip 

\textbf{U.S.A.} ramane indiferentă amenințărilor \textbf{Coreei de Nord}.

\textcolor{green!55!blue}{\textbf{U.S.A.} remains indifferent to \textbf{North Korea}'s threats.}
}
\medskip \hrule \medskip 

\subsection{LOC}

Non-geo-political locations: mountains, seas, lakes, streets, neighbourhoods, addresses, continents, regions that are not GPEs. We include regions such as Middle East, "continents" like Central America or East Europe. Such regions include multiple countries, each with its own government and thus cannot be GPEs.

\medskip \hrule \medskip 

{\footnotesize \fontfamily{qcr}\selectfont
Pe \textbf{DN7 Petroșani-Obârșia Lotrului} carosabilul era umed, acoperit (cca 1 cm) cu zăpadă, iar de la Obârșia Lotrului la stațiunea Vidra, stratul de zăpadă era de 5-6 cm. 

\textcolor{green!55!blue}{On \textbf{DN7 Petroșani-Obârșia Lotrului} the road was wet, covered (about 1cm) with snow, and from Obârșia Lotrului to Vidra resort the snow depth was around 5-6 cm.\footnote{Note: "Obârșia Lotrului" and "Vidra resort" are cities or villages and are thus GPEs; only DN7 which is a national road designation is marked as LOC, including where exacly on DN7 (names of cities are used as markers for the road segment)}}

\medskip \hrule \medskip 

Produsele comercializate în \textbf{Europa de Est} au o calitate inferioară celor din \textbf{vest}.

\textcolor{green!55!blue}{Products sold in \textbf{East Europe} have a lower quality than those sold in the \textbf{west}.\footnote{Note: "west" refers to West Europe and thus we mark it as a LOC.}}

}

\medskip \hrule \medskip 

\subsection{FACILITY}

Buildings, airports, highways, bridges or other functional structures built by humans. Buildings or other structures which house people, such as homes, factories, stadiums, office buildings, prisons, museums, tunnels, train stations, etc., named or not. Everything that falls within the architectural and civil engineering domains should be labeled as a FACILITY. We do not mark structures composed of multiple (and distinct) sub-structures, like a named area that is composed of several buildings, or "micro"-structures such as an apartment (as it a unit of an apartment building). However, larger, named functional structures can still be marked (such as "terminal X" of an airport).

\medskip \hrule \medskip 

{\footnotesize \fontfamily{qcr}\selectfont
\textbf{Autostrada A2} a intrat în reparații pe o bandă, însă pe \textbf{A1} nu au fost încă începute lucrările.

\textcolor{green!55!blue}{Repairs on one lane have commenced on the \textbf{A2 highway}, while on \textbf{A1} no works have started yet.}

\medskip \hrule \medskip 

\textbf{Aeroportul Henri Coandă} ar putea sa fie extins cu un nou \textbf{terminal}.

\textcolor{green!55!blue}{\textbf{Henri Coandă Airport} could be extended with a new \textbf{terminal}.}
}

\medskip \hrule \medskip 

\subsection{PRODUCT}

Objects, cars, food, items, anything that is a product, including software (such as Photoshop, Word, etc.). We don't mark services or processes. With very few exceptions (such as software products), PRODUCT entities have to have physical form, be directly man-made. We don't mark entities such as credit cards, written proofs, etc. We don't include the producer's name unless it's embedded in the name of the product.

\medskip \hrule \medskip 

{\footnotesize \fontfamily{qcr}\selectfont
\textbf{Mașina} cumpărată este o \textbf{Mazda}.

\textcolor{green!55!blue}{The bought \textbf{car} is a \textbf{Mazda}.}

\medskip \hrule \medskip 

S-au cumpărat 5 \textbf{Ford Taurus} și 2 \textbf{autobuze} Volvo.

\textcolor{green!55!blue}{5 \textbf{Ford Taurus} and 2 Volvo \textbf{buses} have been acquired.\footnote{Note: here we won't mark "Volvo" but will mark "Ford" as in one two-word entity "Ford Taurus" as it is embedded in the name.}}

}

\medskip \hrule \medskip 

\subsection{EVENT}

Named events: Storms (e.g.:"Sandy"), battles, wars, sports events, etc. We don't mark sports teams (they are ORGs), matches (e.g. "Steaua-Rapid" will be marked as two separate ORGs even if they refer to a football match between the two teams, but the match is not specific). Events have to be significant, with at least national impact, not local.

\medskip \hrule \medskip 

{\footnotesize \fontfamily{qcr}\selectfont
\textbf{Războiul cel Mare}, \textbf{Războiul Națiunilor}, denumit, în timpul celui de \textbf{Al Doilea Război Mondial}, \textbf{Primul Război Mondial}, a fost un conflict militar de dimensiuni mondiale.

\textcolor{green!55!blue}{The \textbf{Great War}, \textbf{War of the Nations}, as it was called during the \textbf{Second World War}, the \textbf{First World War} was a global-scale military conflict.}

}

\medskip \hrule \medskip 

\subsection{LANGUAGE}

This class represents all languages.

\medskip \hrule \medskip 

{\footnotesize \fontfamily{qcr}\selectfont
Românii din România vorbesc \textbf{română}. 

\textcolor{green!55!blue}{Romanians from Romania speak \textbf{Romanian}.\footnote{Note: we mark  languages, not countries (which are GPEs) or the country's inhabitants (which are NAT\_REL\_POL)}}

\medskip \hrule \medskip 

În Moldova se vorbește \textbf{rusa} și \textbf{româna}.

\textcolor{green!55!blue}{In Moldavia they speak \textbf{Russian} and \textbf{Romanian}.}
}

\medskip \hrule \medskip 

\subsection{WORK\_OF\_ART}

Books, songs, TV shows, pictures; everything that is a work of art/culture created by humans. We mark just their name. We don't mark laws.

\medskip \hrule \medskip 

{\footnotesize \fontfamily{qcr}\selectfont
Accesul la \textbf{Mona Lisa} a fost temporar interzis vizitatorilor.

\textcolor{green!55!blue}{Access to \textbf{Mona Lisa} was temporarily forbidden to visitors.}

\medskip \hrule \medskip 

În această seară la \textbf{Vrei sa Fii Miliardar} vom avea un invitat special.

\textcolor{green!55!blue}{This evening in \textbf{Who Wants To Be A Millionaire} we will have a special guest.}
}

\medskip \hrule \medskip 

\subsection{DATETIME}

Date and time values. We will mark full constructions, not parts, if they refer to the same moment (e.g. a comma separates two distinct DATETIME entities only if they refer to distinct moments). If we have a well specified period (e.g. "between 20-22 hours") we mark it as PERIOD, otherwise less well defined periods are marked as DATETIME (e.g.: "last summer", "September", "Wednesday", "three days"); Ages are marked as DATETIME as well. Prepositions are not included.

\medskip \hrule \medskip 

{\footnotesize \fontfamily{qcr}\selectfont
Te rog să vii aici în cel mult \textbf{o oră}, nu \textbf{mâine} sau \textbf{poimâine}.

\textcolor{green!55!blue}{Please come here in \textbf{one hour} at most, not \textbf{tomorrow} or the \textbf{next day}.}

\medskip \hrule \medskip 

Actul s-a semnat la \textbf{orele 16}.

\textcolor{green!55!blue}{The paper was signed at \textbf{16 hours}.}

\medskip \hrule \medskip 

\textbf{August} este \textbf{o lună} secetoasă.

\textcolor{green!55!blue}{\textbf{August} is a dry \textbf{month}.}

\medskip \hrule \medskip 

Pe \textbf{data de 20 martie} între orele 20-22 va fi oprită alimentarea cu curent.

\textcolor{green!55!blue}{On the \textbf{20th of March}, between 20-22 hours, electricity will be cut-off.\footnote{Note: "20-22 hours" is a PERIOD and not a DATETIME, this is why it is not marked here as such.}}
}

\medskip \hrule \medskip 

\subsection{PERIOD}

Periods/time intervals. Periods have to be very well marked in text. If a period is not like "a-b" then it is a DATETIME.

\medskip \hrule \medskip 

{\footnotesize \fontfamily{qcr}\selectfont
Spectacolul are loc între \textbf{1 și 3 Aprilie}.

\textcolor{green!55!blue}{The show takes place between \textbf{1 and 3 April}.}

\medskip \hrule \medskip 

În prima jumătate a lunii iunie va avea loc evenimentul de două zile.

\textcolor{green!55!blue}{In the first half of June the two-day event will take place.\footnote{Note: "the first half of June" while it is a period, because it is not clearly specified, it will be marked as DATETIME. Also "two-day" is a DATETIME because we don't know exactly which 2 days.}}

}

\medskip \hrule \medskip 

\subsection{MONEY}

Money, monetary values, including units (e.g. USD, \$, RON, lei, francs, pounds, Euro, etc.) written with number or letters. Entities that contain any monetary reference, including measuring units, will be marked as MONEY (e.g. 10\$/sqm, 50 lei per hour). Words that are not clear values will not be marked, such as "an amount of money", "he received a coin".

\medskip \hrule \medskip 

{\footnotesize \fontfamily{qcr}\selectfont
Primarul a semnat un contract în valoare de \textbf{10 milioane lei noi}, echivalentul a aproape \textbf{2.6m EUR}.

\textcolor{green!55!blue}{The mayor signed a contract worth \textbf{10 million new lei}, equivalent of almost \textbf{2.6m EUR}.}
}

\medskip \hrule \medskip 

\subsection{QUANTITY}

Measurements, such as weight, distance, etc. Any type of quantity belongs in this class.

\medskip \hrule \medskip 

{\footnotesize \fontfamily{qcr}\selectfont
Conducătorul auto avea peste \textbf{1g/ml} alcool în sânge, fiind oprit deoarece a fost prins cu peste \textbf{120 km/h} în localitate.

\textcolor{green!55!blue}{The car driver had over \textbf{1g/ml} blood alcohol, and was stopped because he was caught speeding with over \textbf{120km/h} in the city.}
}

\medskip \hrule \medskip 

\subsection{NUMERIC\_VALUE}

Any numeric value (including phone numbers), written with letters or numbers or as percents, which \textit{is not} MONEY, QUANTITY or ORDINAL.

\medskip \hrule \medskip 

{\footnotesize \fontfamily{qcr}\selectfont
Raportul \textbf{XII-2} arată \textbf{4 552} de investitori, iar structura de portofoliu este: cont curent \textbf{0,05\%}, certificate de trezorerie \textbf{66,96\%}, depozite bancare \textbf{13,53\%}, obligațiuni municipale \textbf{19,46\%}.

\textcolor{green!55!blue}{The \textbf{XII-2} report shows \textbf{4 552} investors, and the portfolio structure is: current account \textbf{0,05\%}, treasury bonds \textbf{66,96\%}, bank deposits \textbf{13,53\%}, municipal bonds \textbf{19,46\%}.}
}

\medskip \hrule \medskip 

\subsection{ORDINAL}

The first, the second, last, 30th, etc.; An ordinal must imply an order relation between elements. For example, "second grade" does not involve a direct order relation; it indicates just a succession in grades in a school system.

\medskip \hrule \medskip 

{\footnotesize \fontfamily{qcr}\selectfont
\textbf{Primul} loc a fost ocupat de echipa Germaniei.

\textcolor{green!55!blue}{\textbf{The first} place was won by Germany's team.}
}

\medskip \hrule \medskip 

\medskip \hrule \smallskip \hrule \medskip

\section{Annotation Methodology}

The corpus creation process involved a small number of people that have voluntarily joined the initiative, with the authors of this paper directing the work. Initially, we searched for NER resources in Romanian, and found none. Then we looked at English resources and read the in-depth ACE guide, out of which a 16-class draft  evolved. We then identified a copy-right free text from which we hand-picked sentences to maximize the amount of entities while maintaining style balance. The annotation process was a trial-and-error, with cycles composed of annotation, discussing confusing entities, updating the annotation guide schematic and going through the corpus section again to correct entities following guide changes. The annotation process was done online, in BRAT\footnote{Please note that while the CoNLLU-Plus MWE format supports multi word entities that are not in a \textbf{continuous} sequence, as we performed the annotation in BRAT, we only annotated multi-word contiguous entities.}. 
The actual annotation involved 4 people, has taken about 6 months (as work was volunteer-based, we could not have reached for 100\% time commitment from the people involved), and followed the steps:
\smallskip
\setlist{nolistsep}
\begin{enumerate}[noitemsep]
    \item Each person would annotate the full corpus (this included the cycles of shaping up the annotation guide, and re-annotation). Inter-annotator agreement (ITA) at this point was relatively low, at ~60-70\%, especially for a number of classes.
    \item We then automatically merged all annotations, with the following criterion: if 3 of the 4 annotators agreed on an entity (class\&start-stop), then it would go unchanged; otherwise mark the entity (longest span) as CONFLICTED.
    \item Two teams were created, each with two persons. Each team annotated the full corpus again, starting from the previous step. At this point, class-average ITA has risen to over ~85\%. 
    \item Next, the same automatic merging happened, this time entities remained unchanged if both annotations agreed.
    \item Finally, one of the authors went through the full corpus one more time, correcting disagreements. 
\end{enumerate}
\smallskip

Notes regarding classes and inter-annotator agreements:
\smallskip
\setlist{nolistsep}
\begin{itemize}[noitemsep]
    \item ORGANIZATION, NAT\_REL\_POL, LANGUAGE or GPEs have the highest ITA, over 98\%. 
    \item DATETIME also has a high ITA, with some overlap with PERIOD: annotators could fall-back if they were not sure that an expression was a PERIOD and simply mark it as DATETIME.
    \item WORK\_OF\_ART and EVENTs have caused some problems because the scope could not be properly defined from just one sentence. For example, a fair in a city could be a local event, but could also be a national periodic event.
    \item MONEY, QUANTITY and ORDINAL all are more specific classes than NUMERIC\_VALUE. So, in cases where a numeric value has a unit of measure by it, it should become a QUANTITY, not a NUMERIC\_VALUE. However, this "specificity" has created some confusion between these classes, just like with DATETIME and PERIOD.
    \item The ORDINAL class is a bit ambiguous, because, even though it ranks "higher" than NUMERIC\_VALUE, it is the least diverse, most of the entities following the same patterns. 
    \item PRODUCT and FACILITY classes have the lowest ITA by far (less than 40\% in the first annotation cycle, less than 70\% in the second). We actually considered removing these classes from the annotation process, but to try to mimic the OntoNotes classes as much as possible we decided to keep them in. There were many cases where the annotators disagreed about the scope of words being facilities or products. Even in the ACE guidelines these two classes are not very well "documented" with examples of what is and what is not a PRODUCT or FACILITY. Considering that these classes are, in our opinion, of the lowest importance amongst all others, a lower ITA was accepted.
\end{itemize}
\medskip

Finally, we would like to address the "semantic scope" of the entities - for example, for class PERSON, we do not annotate only proper nouns (NPs) but basically any reference to a person (e.g. through pronouns "she", job position titles, common nouns such as "father", etc.). We do this because we would like a high-coverage corpus, where entities are marked as more semantically-oriented rather than syntactically - in the same way ACE entities are more encompassing than CoNLL entities \footnote{RONEC contains a total of 1251 NPs for the class PERSON. The full list can be found at: \url{https://github.com/dumitrescustefan/ronec/blob/master/ronec/meta/person_proper_nouns.txt}}. 


\section{Conclusions}

We have presented RONEC - the first Named Entity Corpus for the Romanian language. At its current version, in its 5127 sentences we have 26377 annotated entities in 16 different classes. The corpus is based on copy-right free text, and is released as open-source, free to use and extend. There is also an annotation guide that we will improve, and in time evolve into a full annotation document like the ACE Annotation Guidelines for Entities \cite{linguistic2005ace}. 

We have released the corpus in two formats: CoNLL-U PLus (text-based tab-separated pre-tokenized and annotated format, in MWE and IOB flavours) and BRAT (another text-based, non-tokenized format which annotates spans of characters with classes). 

We also release a spaCy\footnote{\hyperlink{https://spacy.io/}{spaCy} is a well-known Python text processing API, offering an easy interface to everything from tokenization to parsing and NER.} pretrained NER model on our GitHub repo for immediate usage.

\section{Bibliographical References}
\label{main:ref}

\bibliographystyle{lrec}
\bibliography{lrec2020W-xample}


\appendix
\section*{Appendix}

\hspace{2pt}

\section{Table Translations}

This appendix contains the translations of the phrases in table \ref{tab:styleexamples} (in order of appearance):

\begin{itemize}
    \item In \textbf{two days}, the  \textbf{Fallujah} battles from \textbf{Iraq} caused the death of \textbf{105 people} and injured more than \textbf{200}.
    \item \textbf{Jean-Claude Juncker}, Prime Minister of \textbf{Luxembourg} was born on \textbf{December 9, 1954}.
    \item \textbf{Tourists} that want to spend their vacations in a quiet, beautiful, and with a minimum of expenses, can opt for spaces specially set up for \textbf{tents} either on the mountain, at the sea or in the \textbf{Danube Delta}.
    \item Also in \textbf{the second} stage, the \textbf{Vardar Skopje} - \textbf{S.C. Pick Szeged} meeting took place, which ended on equal footing, \textbf{24-24}.
    \item Ordinance of the Government no. \textbf{83 / 2004} for amending and supplementing Law no. \textbf{57 / 2003} regarding the Fiscal Code stipulates, at art. \textbf{253}, paragraph (\textbf{6})...
    \item \textbf{SC "Innuendo" S.R.L.} sells in bulk, through its own warehouse located inside the \textbf{"Euro 1" Shopping Center} ...
    \item On \textbf{Cerna Valley} we arrived, this time, on the road (\textbf{DN67D}) from \textbf{Baia de Aramă}.
\end{itemize}

\section{BRAT to CoNLLU-Plus conversion}

The corpus was annotated using BRAT, which does not take into account tokenization: it simply marks character spans with a class. For this reason, the conversion was done by directly tokenizing the text with NLP-Cube and then aligning the resulting tokens to their respective class. There weren't any cases where tokenization produced a token that belonged to 2 different classes. 

\section{SpaCy integration}

At the time of writing the model is not yet integrated natively into spaCy's repo to be able to be used directly. However, we provide a pre-trained model available here\footnote{https://github.com/dumitrescustefan/ronec/tree/master/spacy}. The only extra step is to download this model locally.

Here is how to use spaCy \& RONEC: 

\begin{lstlisting}[language=Python]
import spacy

nlp = spacy.load(<model>)
doc = nlp("Popescu Ion a fost la Cluj.")

for ent in doc.ents:
   print(ent.text, ent.start_char,
	 ent.end_char, ent.label)
\end{lstlisting}

and we should see an output like:

\begin{lstlisting}[language=Python]
Popescu Ion 0 11 PERSON
Cluj 22 26 GPE
\end{lstlisting}

\end{document}